# Kinematic analysis of structural mechanics based on convolutional neural network

Leye Zhang[1], Xiangxiang Tian[1,*], Hongjun Zhang[2]

(1. Jiangsu College of Finance & Accounting, Lianyungang, 222061, China;

2. Wanshi Antecedence Digital Intelligence Traffic Technology Co., Ltd, Nanjing, 210016, China)

**Abstract:** Attempt to use convolutional neural network to achieve kinematic analysis of plane bar structure. Through 3dsMax animation software and OpenCV module, self-build image dataset of geometrically stable system and geometrically unstable system. we construct and train convolutional neural network model based on the TensorFlow and Keras deep learning platform framework. The model achieves 100% accuracy on the training set, validation set, and test set. The accuracy on the additional test set is 93.7%, indicating that convolutional neural network can learn and master the relevant knowledge of kinematic analysis of structural mechanics. In the future, the generalization ability of the model can be improved through the diversity of dataset, which has the potential to surpass human experts for complex structures. Convolutional neural network has certain practical value in the field of kinematic analysis of structural mechanics. Using visualization technology, we reveal how convolutional neural network learns and recognizes structural features. Using pre-trained VGG16 model for feature extraction and fine-tuning, we found that the generalization ability is inferior to the self-built model.

## 0  Introduction

Any engineering building (bridge, house, etc.) has structural mechanics problems to be solved. The general structure is actually a spatial structure, but in most cases, it can be decomposed into a plane structure, so that the calculation can be simplified. The structures are the composition of some rods, plates, shells, entities, etc. A structure must be able to bear the load, first of all, its own geometry must be able to remain unchanged. So in the engineering design stage, it is necessary to carry out kinematic analysis (analysis of geometrical construction). Kinematic analysis of plane bar structure is the most basic content in structural mechanics[1].

In recent years, artificial intelligence technology has made remarkable progress and gradually penetrated into various industries, including the field of engineering design. Cheng Guozhong proposed an intelligent design method for high-rise shear wall structures based on deep reinforcement learning[2]; Lu Xinzheng proposed an intelligent design optimization method for shear walls based on rule learning and coding[3]; Zheng Zhe proposed a response prediction model of structural mechanics based on graph neural network[4]; Zhang Hongjun used a denoising diffusion implicit model to generate new bridge types[5]; Wu Rih-Teng estimated the dynamic response of steel frames based on convolutional neural network[6]; Stoffel Marcus used convolutional neural network to predict the structural deformation of metal plates under impact loads[7]. However, the application of artificial intelligence technology in kinematic analysis of structural mechanics has not yet been explored by scholars.

Convolutional Neural Network (CNN) is a widely used deep learning model in computer vision applications. In this paper, we build a self-built image dataset of geometrically stable system and geometrically unstable system, and use convolutional neural network to analyze kinematic analysis of plane bar structure (open source address of this article's dataset and source code :https://github.com/

zhangleye/Kinematic-Analysis).

# 1 Introduction to kinematic analysis of structural mechanics and convolutional neural network

## 1.1 Kinematic analysis of structural mechanics

For the sake of convenience, only the plane bar structure is discussed here, and the support and foundation are not considered. The deformation of the material is not taken into account during the analysis.

1. Geometrically stable system: The geometry of the structure itself remains unchanged under arbitrary loads, such as the articulated triangle in Figure 1.

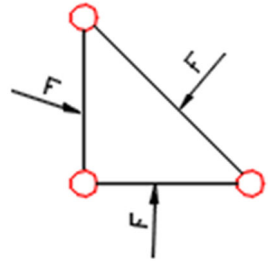

Fig.1 Example of geometrically stable system

2. Geometrically unstable system: The geometry of the structure itself cannot be maintained under very small loads, such as the articulated quadrilateral in Figure 2.

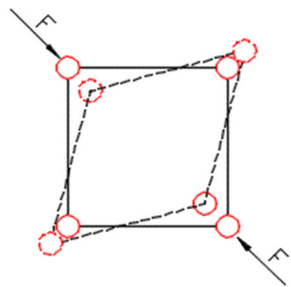

Fig.2 Example of geometrically unstable system

The black line in Fig. 1 and Fig. 2 indicates the bar, the red circle represents the hinge joint, the F represents the external load, and the dotted line indicates the geometry at a certain point in time when the mechanical movement occurs.

The general engineering structure must be a geometrically stable system, and a geometrically unstale system cannot be adopted, otherwise the balance cannot be maintained under the load.

## 1.2 Convolutional neural network

The study of convolutional neural network began in the 80s and 90s of the twentieth century. With the development of deep learning theory and massively parallel chips (graphics processors, Gpus), convolutional neural network has been widely used in computer vision and other fields. In the 2012 American movie "Act of Valor", which is known for its realism, the aerial reconnaissance drone identifies the activities of ground personnel and vehicles in real time, and the core algorithm is a convolutional neural network.

Convolutional neural network is built by mimicking the visual perception mechanism of living organisms. The fundamental difference between a dense layer and a convolutional layer (Convnet) is that the dense layer learns a global pattern from the input feature space, while the convolutional layer learns a local pattern. This feature gives the patterns learned by the convolutional neural network a translation invariant. After a convolutional neural network learns a pattern in the lower right corner of an image, it can recognize it anywhere, such as in the upper left corner. For a fully-connected network, if a pattern appears in a new location, it can only relearn the pattern. This allows convolutional neural network to make efficient use of data when processing images, and it requires fewer training samples to learn data representations with generalization capabilities. At the same time, by downsampling the feature map, the observation window of the convolutional layer is larger and larger (relative to the original input image), so that the convolutional neural network can learn the spatial hierarchies of patterns, which makes the convolutional neural network can effectively learn abstract visual concepts [8].

Convolutional neural network for image classification generally consists of two parts: a convolutional base (composed of multiple convolutional layers) and a classifier (generally composed of two dense layers). The convolutional base is responsible for feature extraction, and the classifier is responsible for categorical probability coding.

## 2 Kinematic analysis of structural mechanics based on convolutional neural network

### 2.1 Task description

A convolutional neural network model is established for image classification of geometrically stable system and geometrically unstable system, so as to realize kinematic analysis of plane bar structure. The basic grasp of the model is required: (1) The hinge joint can rotate relatively, but the rigid joint cannot rotate relatively; (2) Adding or subtracting an unit of two components, stability of the system do not change; (3) The hinged triangle is stable, and the hinged quadrilateral is unstable.

A completely random baseline (dumb baseline) is 50% accuracy. For all samples in this article, the human expert's baseline is 100% accuracy because they are simple structures. (Note: For complex structures, human experts cannot directly identify the stability, and they need to be solved by means of computer analysis or other.) Therefore, convolutional neural network model has the condition to surpass human experts in kinematic analysis of complex structures, but they are not covered in this paper. )

### 2.2 Self-built dataset

Although there are some structural examples in structural mechanics textbooks, it is difficult to use them as dataset. Therefore, 3dsMax animation software and OpenCV module are used to build own dataset. Due to cost constraint, there are only 12 structural examples of geometrically stable system and geometrically unstable system, as shown in Fig. 3 and Fig. 4. (In the diagram, if two bars are directly intersected, they are rigid connected.)

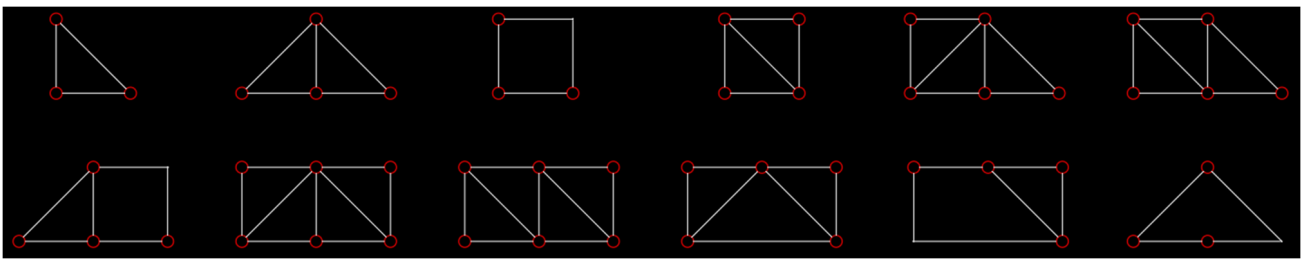

Fig.3 Twelve structural examples of geometrically stable system in dataset

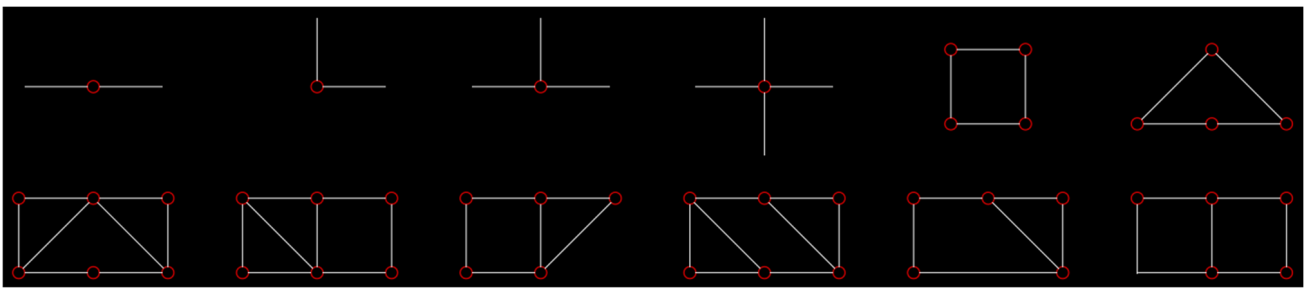

Fig.4 Twelve structural examples of geometrically unstable system in dataset

The specific steps are as follows: (1) Draw structure examples in 3dsMax, use the camera position animation from near to far, and render 9 color pictures (Fig. 5) for each structure example, 256x256 pixels, png format; (2) Then use the OpenCV module to rotate each picture, and the rotation angle is 0, 40, 80, 120, 160, 200, 240, 280, 320 degrees (Figure 6) ;(3) Finally, the OpenCV module is used to translate each picture, horizontally and Vertically translate plus or minus 10 pixels for a total of 9 combinations (Figure 7). (Note: Scaling, rotating, and translating are essentially just data augmentations.)

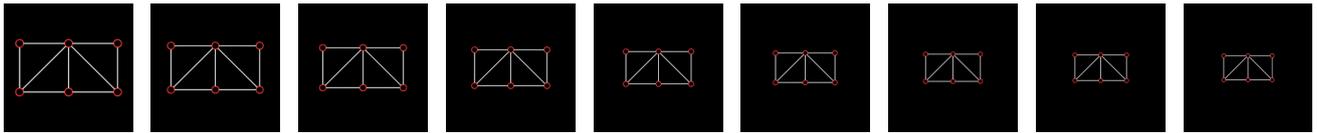

Fig.5 Nine results of 3dsMax animation rendering for a certain structural example

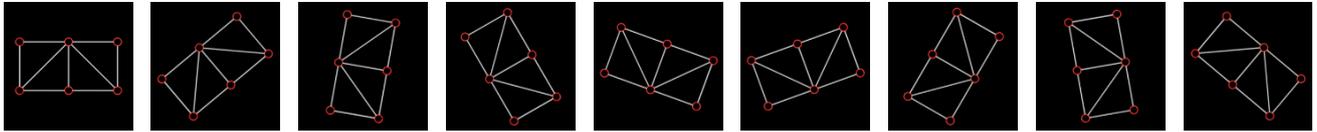

Fig.6 Nine results of OpenCV rotation operation for a certain image

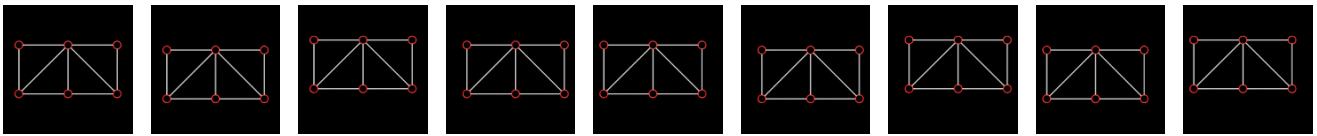

Fig.7 Nine results of OpenCV translation operation for a certain image

Each structure example has 9x9x9=729 different images, and the entire dataset has a total of 729x24=17496 images. The dataset is randomly scrambled, and the dataset is divided into training set (50%), validation set (25%), and test set (25%).

## 2.3 Construction and training of convolutional neural network

This task is a problem of binary classification. Based on the Python 3.10 programming language, TensorFlow 2.10 and Keras2.10 deep learning platform frameworks, convolutional neural network is built and trained.

1. Architecture of model

The convolutional base is stacked with 6 sets of Conv2D, MaxPooling2D, and Dropout layers. The Conv2D layer has a sample window size of 3x3 and an activation function of relu, and the MaxPooling2D layer has a window size of 2x2 and a dropout layer ratio of 0.2. The number of filters in the Conv2D layer is 4, 4, 8, 8, 16, and 16.

The classifier is composed of two dense layers, the number of neurons is 16, 1, and the activation function is relu and sigmoid.

Tab.1 Model summary of convolutional neural network

| Layer (type) | Output Shape | Param # |
|---|---|---|
| conv2d(Conv2D) | (None,256,256,4) | 112 |
| max_pooling2d(MaxPooling2D) | (None,128,128,4) | 0 |
| dropout(Dropout) | (None,128,128,4) | 0 |
| conv2d_1(Conv2D) | (None,128,128,4) | 148 |
| max_pooling2d_1(MaxPooling2D) | (None,64,64,4) | 0 |
| dropout_1(Dropout) | (None,64,64,4) | 0 |
| conv2d_2(Conv2D) | (None,64,64,8) | 296 |
| max_pooling2d_2(MaxPooling2D) | (None,32,32,8) | 0 |
| dropout_2(Dropout) | (None,32,32,8) | 0 |
| conv2d_3(Conv2D) | (None,32,32,8) | 584 |
| max_pooling2d_3(MaxPooling2D) | (None,16,16,8) | 0 |
| dropout_3(Dropout) | (None,16,16,8) | 0 |
| conv2d_4(Conv2D) | (None,16,16,16) | 1168 |
| max_pooling2d_4(MaxPooling2D) | (None,8,8,16) | 0 |
| dropout_4(Dropout) | (None,8,8,16) | 0 |
| conv2d_5(Conv2D) | (None,8,8,16) | 2320 |
| max_pooling2d_5(MaxPooling2D) | (None,4,4,16) | 0 |
| dropout_5(Dropout) | (None,4,4,16) | 0 |
| flatten(Flatten) | (None,256) | 0 |
| dense(Dense) | (None,16) | 4112 |
| dense_1(Dense) | (None,1) | 17 |
| Total params: 8,757 | | |
| Trainable params: 8,757 | | |
| Non-trainable params: 0 | | |

2. Loss Function

For the binary classification problem, the model output is a scalar in the range of 0~1 (the probability value labeled 1), and the loss function adopts binary cross-entropy.

$$Loss = 1/n * \sum \{y*\ln(1/y') + (1-y)*\ln[1/(1-y')]\}$$

In this formula: Loss is the binary cross-entropy loss; y is the sample label; y' is the probability value of the model predicting the sample label is 1; n is the number of samples; To avoid the denominator being 0, the denominator needs to be added with a tiny value (2e-07), which is not expressed in the formula.

Calculation example: y=[0, 0, 1, 1], y'=[0.02, 0.03, 0.99, 0.0.97], then Loss=1/4*(0.020+0.030+0.010+0.030)=0.023. Accuracy (the proportion of correct classification) = (1+1+1+1)/4=100%.

3. Training

The "Adam" optimizer is used to update the neural network parameters, and the monitoring indicator is "accuracy". The Python generator continuously reads the images for the model to call. Use the callback function to save the model weights for each epoch.

The loss curve of the training process is shown in the figure below (the first 100 epochs):

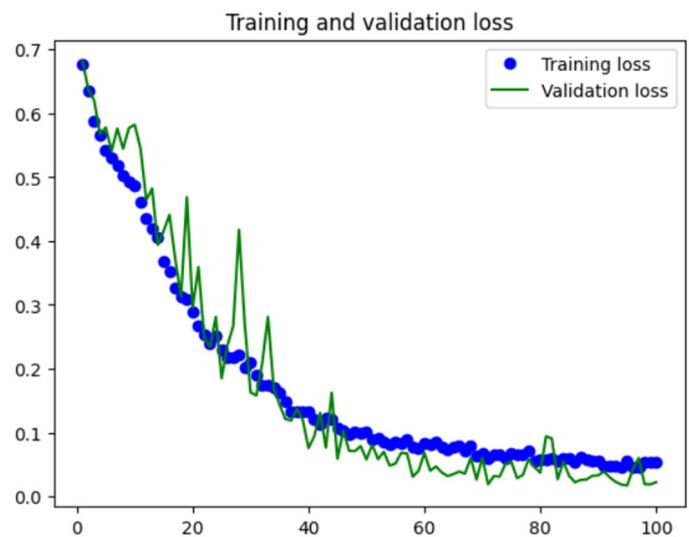

Fig.8 Training loss curve

The accuracy curve of the training process is shown in the following figure (the first 100 epochs):

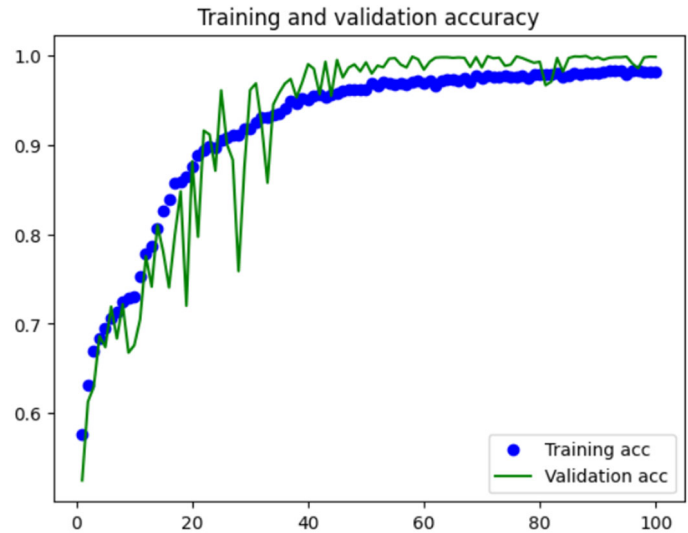

Fig.9 Training accuracy curve

After 1000 epochs of training, the accuracy of the model in the training set, validation set, and test set is 100%, and the model performance is satisfactory. However, it should be noted that the structural examples in the test set have all been seen by the model in the training set, and only the parameters of scaling, rotation, and translation are different. Therefore, the generalization performance of the model needs to be further tested.

## 2.4 Additional test

The goal of machine learning is to get models that can be generalized, that is, models that perform well on data that has never been seen before.

After the model training is completed, 10 new examples of completely different structures are created, as shown in Figure 10.

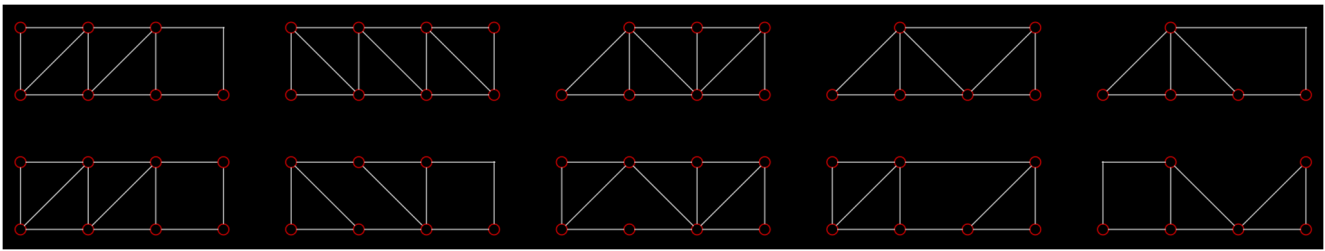

Fig.10 Ten additional completely different structural examples (the first row is geometrically stable system, the second row is geometrically unstable system)

By scaling, rotation, and translation, each structure example has 9x9x9=729 different images, and the additional test set has a total of 729x10=7290 images. The accuracy of the model on this additional test set is 93.7%.

It can be seen that the model with only 8757 parameters has good generalization ability and has mastered some ability of kinematic analysis of structural mechanics, but it cannot reach the baseline of human experts. This is due to the fact that there are only 24 structural examples in the dataset, and it is difficult for the model to fully learn kinematic analysis of structural mechanics with such little knowledge. The solution is to greatly increase the number of dataset structure examples, which can be automatically generated with the help of generative deep learning technology, so as to reduce the workload of manual dataset production and realize self-iterative update of the model.

## 3  Explain what convolutional neural network have learned

It is often said that deep learning models are "black boxes" and that the representations learned by the models are difficult to extract and present in a way that humans can understand. Convolutional neural network, on the other hand, learns representations that can be visualized because they are representations of visual concepts. Here we try three techniques [8], invoking the model saved in the previous section, visualizing what the convolutional neural network learns, and helping to understand the decisions made by the neural network.

## 3.1 Visualizing intermediate activations

Visualizing intermediate activations is the representation of the feature maps of the output of each convolutional layer for a given input, which helps to understand how the convolutional layer transforms the input.

Here, an image of the test set is input into the model, and then the feature map output by the first convolutional layer of the model is plotted as a two-dimensional image (4 images are stitched horizontally, see Figure 11).

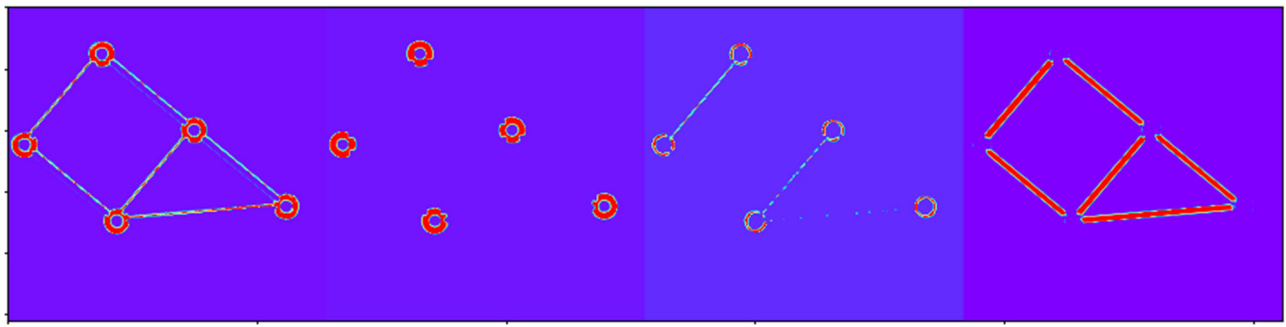

Fig.11 Output feature maps of the first convolution layer

It is clear that the 2nd channel should be a hinge detector and the 4th channel should be a bar detector.

## 3.2 Visualizing convnet filters

Convolutional neurons act as filters, a series of mixed wave inputs of different frequencies, and the filter prefers to amplify wave of a specific frequency, i.e., the filter responds most to a particular wave. By analyzing the wave with the largest response of the filter, we can know the function and mode of the filter.

Visualizing the filters of a convolutional neural network helps to accurately understand the visual pattern of each filter response in a convolutional neural network. This can be achieved by gradient ascent in the input space: starting with a random input image, the gradient ascent is applied to the value of the convolutional neural network input image, with the aim of maximizing the response of a certain filter, and the updated input image is the image with the largest response of that filter.

Here a random image (Figure 12 far left image) is set up, fed to the individual neurons of the first convolutional layer, and then gradient ascending to maximize filter activation as follows (Figure 12 right 4 images):

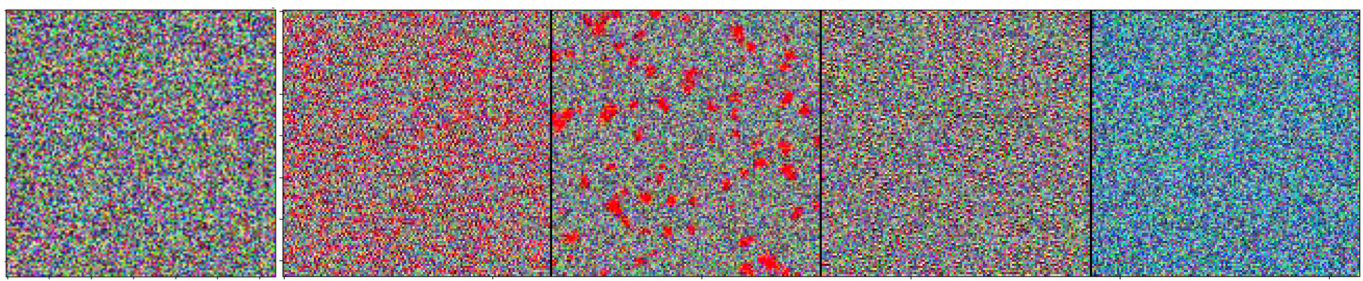

Fig.12 Each filter mode of the first convolution layer

The second channel, which responds most to the red plaque signal, should be used to extract the information from the hinge.

## 3.3 Visualizing heatmaps of class activation

A "class activation" heatmap is a 2D grid of scores associated with an specific output class, computed for every location in any input image, indicating how important each location is with respect to the class considered. It helps to understand which parts of a given image led a convnet to its final classification decision. It consists in taking the output feature map of a convolution layer given an input image, and weighing every channel in that feature map by the gradient of the class with respect to the channel.

Here, an image of the test set is input into the model, the feature map of the last convolutional layer of the model is output, the gradient of the category relative to the feature map is obtained, and the class activation heat map is weighted. Overlay the heatmap on top of the original image, as shown below:

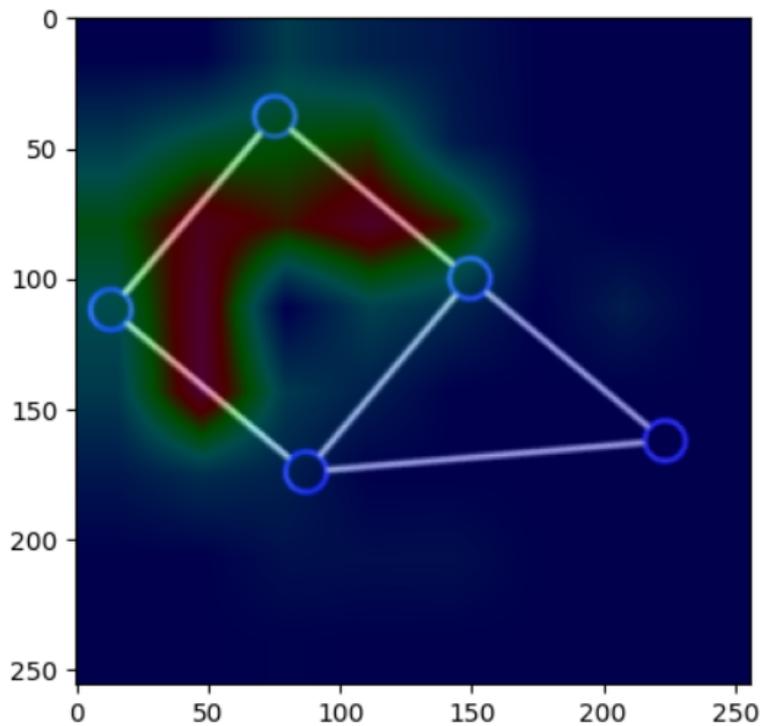

Fig.13 Superimposing the class activation heatmap with the original picture

It can be seen that it is the hinged quadrilateral that allows the convolutional neural network to make the decision of the geometrically unstable system.

## 4 Using a pre-trained convnet

A pre-trained network is simply a saved network previously trained on a large dataset, typically on a large-scale image classification task. If this original dataset is large enough and general enough, then the spatial feature hierarchy learned by the pre-trained network can effectively act as a generic model of our visual world, and hence its features can prove useful for many different computer vision problems, even though these new problems might involve completely different classes from those of the original task[8].

A large convolutional neural network VGG16 trained on the ImageNet dataset is used. ImageNet has 1.4 million labeled images and 1,000 different classes, including buildings, landscapes, animals, people, airplanes, and more.

There are two ways to use a pretrained model: feature extraction and fine-tuning the model.

### 4.1 Feature extraction

Feature extraction is to take the previously trained convolutional base, run new task data on it, and then train a new classifier on top of the output.

Start by instantiating the VGG16 convolutional base, and then add a fully-connected classifier. The classifier is composed of two dense layers, the number of neurons is 128, 1, and the activation functions are relu and sigmoid.

After 100 epochs of training (the convolutional base is frozen), the accuracy of the model is 100% on the training set, the validation set, and the test set, and 67.3% on the additional test set, which shows that the generalization ability is not ideal, which should be related to the large difference between kinematic analysis of structural mechanics and the conventional visual task.

### 4.2 Fine-tuning the model

Fine-tuning consists in unfreezing a few of the top layers of a frozen model base used for feature extraction, and jointly training both the newly added part of the model (in our case, the fully-connected classifier) and these top layers. This is called "fine-tuning" because it slightly adjusts the more abstract representations of the model being reused, in order to make them more relevant for the problem at hand.

The block5_conv2 and block5_conv3 of the VGG16 convolutional base were unfrozen, and the other layers of the convolutional base were frozen, and then the two unfrozen layers and the classifier were jointly trained for 50 epochs, and the accuracy of the model was 100% on the training set, verification set, and test set, and 89.2% on the additional test set. It can be seen that the generalization ability of the model has been greatly improved after fine-tuning, but it is still inferior to the author's self-built model.

## 5 Conclusion

Through the self-built image dataset and convolutional neural network, kinematic analysis of structural mechanics of the plane bar structure is successfully realized. Although the accuracy on the additional test set is not at the level of human experts, the accuracy rate of 93.7% has shown that the model has good generalization ability. In the future, expanding the size and diversity of the dataset will help improve the generalization ability of the model. It is expected that convolutional neural network will have the ability to surpass human experts in the face of complex structures, and they will have certain practical value in the field of kinematic analysis of structural mechanics. This study provides a new perspective and method for the field of kinematic analysis of structural mechanics.

Visualization techniques are used to explore how convolutional neural network learns and identifies structural features. By comparing the pre-trained VGG16 model, we found that the self-built model has better generalization ability for specific tasks.